\documentclass[10pt,twocolumn,letterpaper]{article}

\usepackage[pagenumbers]{cvpr} 

\pdfoutput=1
\usepackage{multirow}
\usepackage{adjustbox}
\usepackage{color}
\usepackage{amsmath,amssymb,amsfonts}
\usepackage{CJKutf8}
\usepackage[hyphens]{url}  
\usepackage{graphicx} 
\usepackage{bbm}

\usepackage[dvipsnames]{xcolor}

\definecolor{cvprblue}{rgb}{0.21,0.49,0.74}
\usepackage[pagebackref,breaklinks,colorlinks,citecolor=cvprblue]{hyperref}

\def\paperID{14565} 
\def\confName{CVPR}
\def\confYear{2024}

\title{MCF-VC: Mitigate Catastrophic Forgetting in Class-Incremental Learning for Multimodal Video Captioning}

\author{Huiyu Xiong\and Lanxiao Wang\and Heqian Qiu\and Taijin Zhao\and Benliu Qiu\and Hongliang Li \\
University of Electronic Science and Technology of China, Chengdu, China\\
{\tt\small \{hyxiong, lanxiao.wang, zhtjww, qbenliu\}@std.uestc.edu.cn}
{\tt\small \{hqqiu, hlli\}@uestc.edu.cn}
}

\begin{document}
\begin{CJK}{UTF8}{gbsn}
\maketitle
\begin{abstract}
To address the problem of catastrophic forgetting due to the invisibility of old categories in sequential input, existing work based on relatively simple categorization tasks has made some progress. In contrast, video captioning is a more complex task in multimodal scenario, which has not been explored in the field of incremental learning. 
After identifying this stability-plasticity problem when analyzing video with sequential input, we originally propose a method to \textbf{M}itigate \textbf{C}atastrophic \textbf{F}orgetting in class-incremental learning for multimodal \textbf{V}ideo \textbf{C}aptioning (MCF-VC). As for effectively maintaining good performance on old tasks at the macro level, we design \textbf{F}ine-\textbf{g}rained \textbf{S}ensitivity \textbf{S}election (FgSS) based on the Mask of Linear’s Parameters and Fisher Sensitivity to pick useful knowledge from old tasks. Further, in order to better constrain the knowledge characteristics of old and new tasks at the specific feature level, we have created the \textbf{T}wo-\textbf{s}tage \textbf{K}nowledge \textbf{D}istillation (TsKD), which is able to learn the new task well while weighing the old task. Specifically, we design two distillation losses, which constrain the cross modal semantic information of semantic attention feature map and the textual information of the final outputs respectively, so that the inter-model and intra-model stylized knowledge of the old class is retained while learning the new class. In order to illustrate the ability of our model to resist forgetting, we designed a metric $\widetilde{\mathcal{CIDER}}_t$ to detect the stage forgetting rate. Our experiments on the public dataset MSR-VTT show that the proposed method significantly resists the forgetting of previous tasks without replaying old samples, and performs well on the new task.
\end{abstract}    
\section{Introduction}
\label{sec:intro}

\begin{figure}[t] 
\centerline{\includegraphics[scale=0.41]{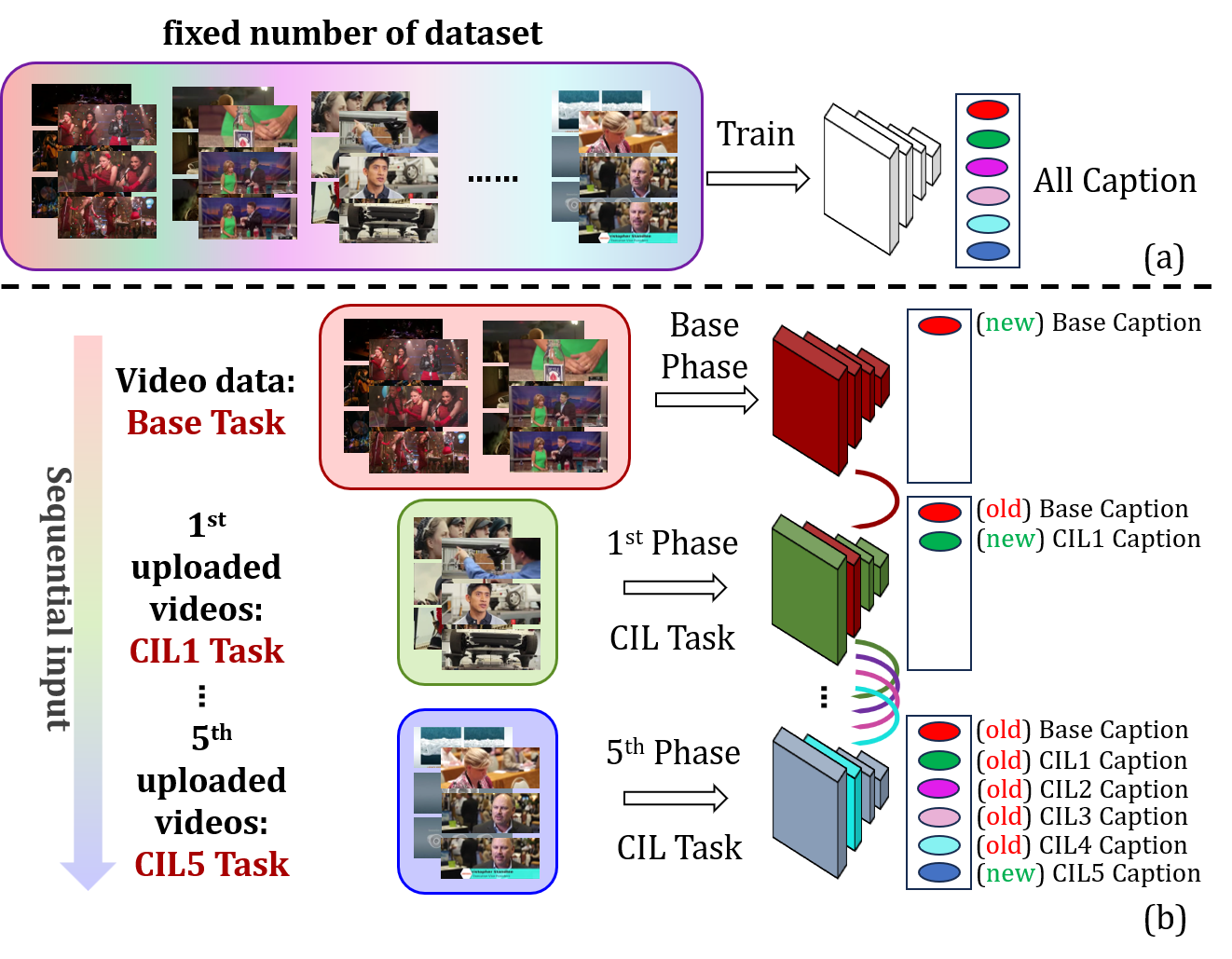}}
\vspace{-0.5cm} 
\caption{Comparison of traditional video captioning and incremental video captioning. In traditional video captioning (a), all visual-language pairs are obtained during training. However, in incremental tasks (b), new class orders arrive while none of the old data is visible. Therefore, the model needs to be updated to learn the new class while maintaining the accuracy of the old class.
}
\vspace{-0.6cm} 
\label{fig1}
\end{figure}

As a branch of video understanding and analysis~\cite{2018Attend, 2019Spatio, falcon2022feature}, video captioning is a typical example of exploring computer intelligence that integrates visual perception and linguistic expression synergistically~\cite{inproceedings} with the aim of automatically describing the visual content of video in natural language. The amount of videos published to the Internet is growing daily due to the quick development of social media such as Facebook and Instagram. Consequently, it becomes imperative to be able to reliably summarize the massive volume of video data that is always being added. The traditional works shown in Figure~\ref{fig1}(a) have made some progress by feeding the fixed number of dataset during training. However, with new video data constantly emerging in the real world, the most intuitive solution is mixing all videos of old and new tasks and training a new model from scratch. Nevertheless, when a new category appears, retraining the model every time is not feasible due to copyright restrictions, storage difficulties of videos and time-consuming. Directly fine-tuning the new sequential task based on the old model can be the serious catastrophic forgetting~\cite{McCloskey1989Catastrophic,1995Catastrophic,2013An}, given the fact that old data is not visible. To solve this problem, training a high-performance model (shown in Figure~\ref{fig1} (b)) that can understand new videos without forgetting knowledge from old videos is a meaningful research direction.

As a newly proposed problem, class-incremental video captioning has a certain correlation with the class-incremental learning task on image classification. Solving the catastrophic forgetting in classification can be mainly divided into replay, dynamic expansion and knowledge distillation~\cite{de2021continual}. Previous work involved knowledge transfer only visual features without cross-modal features, so class-incremental learning for classification is hardly applied to more complex video captioning. In a nutshell, it is significant to design a unique incremental model that can be applied to video captioning tasks.

Considering the above problems, this paper proposes the model called \textbf{M}itigate \textbf{C}atastrophic \textbf{F}orgetting in class-incremental learning for multimodal \textbf{V}ideo \textbf{C}aptioning (MCF-VC). First, we fine-tune the backbone by adding structured dropout and Glossary Ensemble to improve the whole structure that is high-performance and suitable for receiving sequential input data. According to the order of ``screening"-``inheritance"-``optimization", we design two modules for class-incremental learning. We design \textbf{F}ine-\textbf{g}rained \textbf{S}ensitivity \textbf{S}election (FgSS) to screen and inherit useful knowledge from old classes for new class training. The design of this part fully takes into account the characteristics of the encoder-decoder architecture for video captioning, using fisher sensitivity to screen the parameter's gradient that has been pruned by the linear masks and the important frozen layers to inherit some critical parameters of the old model. During optimization, \textbf{T}wo-\textbf{s}tage \textbf{K}nowledge \textbf{D}istillation(TsKD) is used to constrain the cross-modal features of the intermediate layer and the final output text. Specifically, in the first stage, we grasped the cross-modal features of the middle layer and used stylized constraints to ensure that the old style is migrated to the new features. After comparing the differences between the two text outputs in the second stage, we mainly hope to bring the distribution between this two outputs closer to achieve the purpose of balancing the old and new information.
In general, the key contributions of this paper can be summarized as follows:
\begin{itemize}
    \item As the first work to explore the incremental learning on video captioning, we propose MCF-VC method for the special structure of the video captioning backbone to deal with the problems of forgetting the old task caused by sequential input and the poor performance of the new task.
    \item In response to the particularity of the incremental video captioning task, we modify the backbone network to make it more suitable for accepting subtasks with continuous inputs.
    \item In order to make full use of the useful knowledge in the old models, we design a selector to choose information adaptively from fine-grained old trained parameters that is valuable to new tasks, thereby improving the performance of the target model.
    \item For the output of different stages of the backbone network, we propose two different ways of distillation, which are used to restrain the representations of cross-modal and text information, so that the model can balance new and old knowledge.
    \item Some quantitative and qualitative experiments on the public dataset MSR-VTT show that the incremental method proposed in this paper has significant performance improvements on both martics designed in this paper for evaluating class-incremental video captioning oblivion and martics in natural language processing.
\end{itemize}

\section{Related Work}

This paper proposes an original work, that is, the incremental video captioning under sequential task settings. Traditional works of video captioning address the problem of video comprehension on video datasets that are primarily targeted at fixed data volumes. Next, based on the incremental learning setting, we investigate some existing worthy classical methods.

\begin{figure*}[htbp]
\centerline{\includegraphics[scale=0.48]{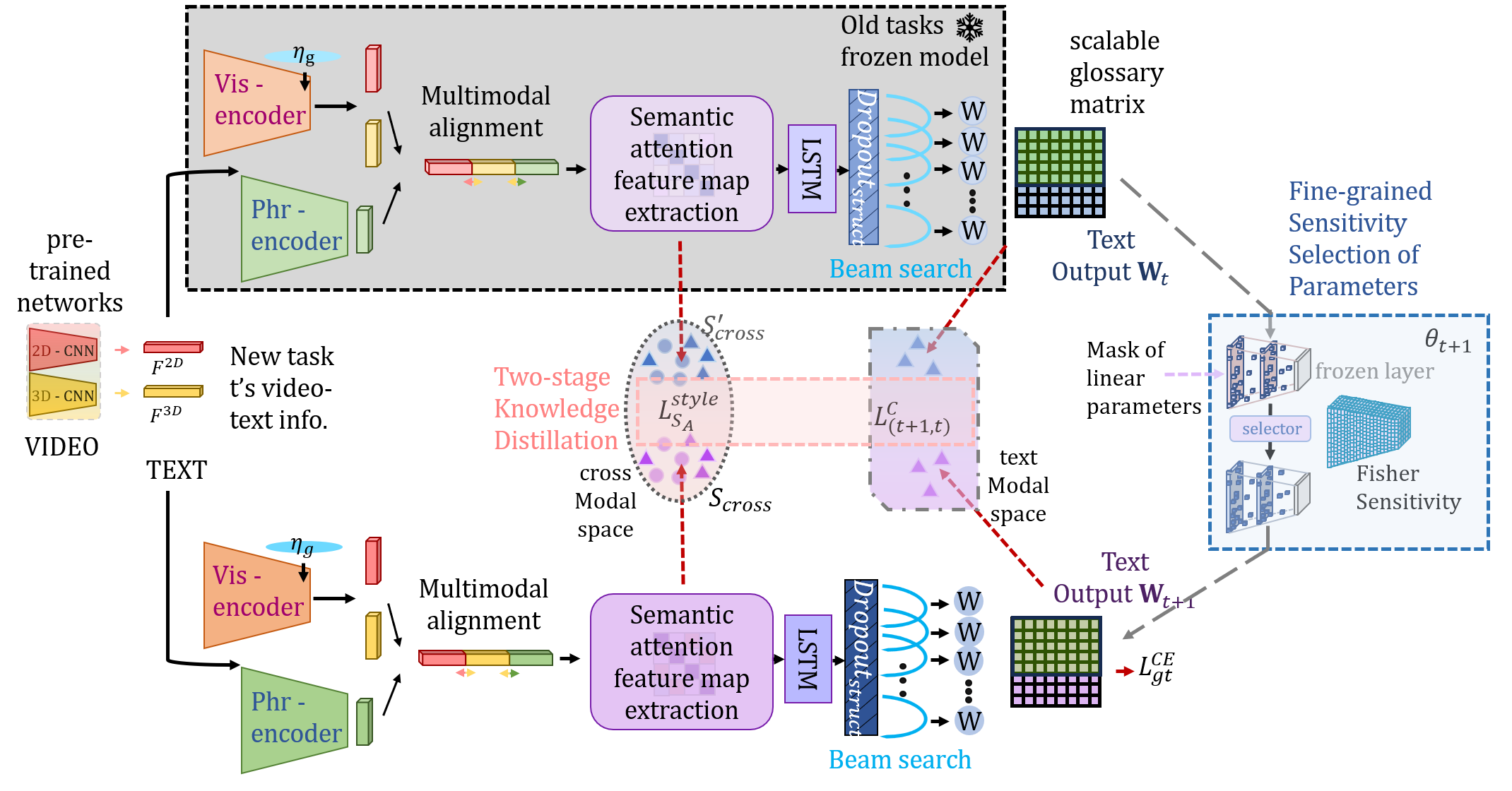}}
\vspace{-0.3cm} 
\caption{The graphical illustration of our approach MCF-VC for the new class-incremental task. Enter the dynamic ${\mathcal{F}}^{3D}$ and static $\mathcal{F}^{2D}$ visual features of the new task and their corresponding text from the data pre-processing into a modified video caption backbone. In order to be more suitable for the task of class-increment, we design FgSS, which masks the fine-grained gradient information in three steps, so that the new model can obtain a balanced effect on the new and old tasks. Next, the cross-modal feature map $(S_A, S'_A)$ extracted by semantic attention and the final output text features $(\mathbf{W}_{t+1}, \mathbf{W}_{t})$ will undergo TsKD.
}
\vspace{-0.3cm} 
\label{fig2}
\end{figure*}

\paragraph{Video Captioning}

Early methods of video caption~\cite{2002Natural, guadarrama2013youtube2text} are largely manual. Thanks to the development of deep learning, in recent years more and more methods has been based on encoder-decoder~\cite{venugopalan2014translating} structures. Heavily depend on the characteristics of RNN, this structure processes the video features in two phases to obtain coherent statements. In order to fully tap into the rich temporal and spatial information present in the video, some methods~\cite{chen2018less, ryu2021semantic, xiong2022stsi} choose to use attention-based modules. At the same time, noting the peculiarities of sentence components, some research~\cite{pan2020spatio, zou2021maps, wang2021pos, li2022graph} has begun to focus on the semantics and part-of-speech coherence of generative descriptions. In recent years, with the boom of large-scale models~\cite{yang2023vid2seq, maaz2023video, wu2023cap4video, long2023spring}, solving such problems has gradually begun to rely on the rich prior knowledge provided by web-scale pre-trained models. Unlike them, we aim to solve the forgetfulness that comes with the sequential input. Therefore, we improve the transformer-based pipeline by adding some modules. While improving traditional performance, it can also settle class-incremental task.

\paragraph{Incremental Learning}

The purpose of incremental learning is to continuously learn useful knowledge and optimize the model by accumulating a series of data~\cite{liu2021adaptive} that appear periodically. Existing work can be broadly divided into three branches: replay, dynamic expansion and knowledge distillation. 
Replay method adds a small amount of old data to the process of new training. In some ways, researchers~\cite{rebuffi2017icarl, castro2018end} try to select a certain percentage of representative samples from old classes to participate in subsequent training. On the other hand, some works~\cite{1995Catastrophic, pourkeshavarzi2021looking, yu2023overcoming} synthesize the distributions of old data as prior knowledge for joint training.
Dynamic expansion method uses a progressive neural network to expand the structure~\cite{aljundi2017expert, yoon2017lifelong, rosenfeld2018incremental, 2021DER} as new tasks arise, so that the knowledge of old tasks can be inherited from the previously learned network.
Distilled method will design the regularization term~\cite{li2017Learning, kirkpatrick2017overcoming} in the loss section to update the whole network. This term is added between the old and new networks, forcing the characteristics~\cite{liu2020mnemonics} of new model to remain partially close to the old one.
Inspired by the above research, our work specifically focuses on the characteristics of the video captioning backbone network, to find a solution for ``data-free".




\section{Methodology}

In order to address the sequential input issue in challenging video comprehension tasks, we create an incremental learning framework (As shown in Figure~\ref{fig2}) for the generation of video captions. We first carry out a preliminary problem setup for this new assignment by constructing a video captioning backbone. Next, various incremental techniques based on that backbone are investigated in detail.

\subsection{Preliminary: Problem Setting}

Class-incremental video captioning aims to analyze captioning content in a sequence of tasks containing an increasing number of video categories. Specifically, we assume that there exists a sequential tasks $\mathbf{T}={\{ t\}}^{\mathcal{T}}_{t=1}$ with the number of tasks $\mathcal{T}$, where $t$ is the $t$-th task. The entire data $\mathbf{D}$ in this sequence consists of the video set ${\{ \mathbf{V}_t\}}^{\mathcal{T}}_{t=1}$ and the manual annotation set ${\{ \mathbf{W}^*_t\}}^{\mathcal{T}}_{t=1}$. It can also be denoted as $\mathbf{D}={\{ \mathbf{D}_t\}}^{\mathcal{T}}_{t=1}={\{ (\mathbf{V}_t, \mathbf{W}^*_t)\}}^{\mathcal{T}}_{t=1}$, where $\mathbf{D}_t$ denotes the data samples of the task $t$.

In order to make the model realize the generation of descriptive discourse that most closely matches the input video, we aim to train a function to satisfy
$f:\mathbf{V} \rightarrow \mathbf{W}^*$. We assume that the model has learned t tasks. When confronted with the new task $t+1$ and the corresponding data $\mathbf{D}_{t+1}$, we aim to update the knowledge learned by the model while utilizing the help of the prior knowledge learned from the old tasks. This process requires the model to be able to find parameters minimizing the following loss function in the current task:

\begin{equation}
min \sum^{\mathcal{T}}_{t=1}
\mathbb{E}_{\mathbf{D}_{t+1}}[\mathcal{L}_{total}(f_{t+1}(\mathbf{V}_{t+1}, \theta_{t+1}|\Theta_{t}), \mathbf{W}^*_{t+1})]
\label{eq13}
\end{equation}

where $\mathcal{L}_{total}$ represents the total loss function of the model, while $\Theta_{t}$ and $\theta_{t+1}$ denote the corresponding parameters of the model after learning the old task and the new task, respectively.

\subsection{Process Multimodal Information to Generate Captions}

As a newly introduced task, the backbone adopted for class-incremental video captioning requires the following two aspects. On the one hand, this adapted network can improve the performance on independent tasks. On the other hand, it helps the network to memorize old knowledge and facilitates the addition of incremental learning strategies.

\paragraph{Backbone design for Appropriate Captioning}
Inspired by the SGN~\cite{ryu2021semantic}, the main framework of our network for video captions generation relies on encoder-decoder. For the input video $V$, we sample $\ell$ frames and $\ell$ clips. And the $i$th frame is ${\mathbf{f}}_{i=1}^{\ell}$ In the same way, the $i$th clip is ${\mathbf{c}}_{i=1}^{\ell}$. The static feature ${\mathcal{F}}^{2D}$ and dynamic feature ${\mathcal{F}}^{3D}$ are extracted from ${\mathbf{f}}_{i=1}^{\ell}$ and ${\mathbf{c}}_{i=1}^{\ell}$ using the pretrained 2D-CNN and 3D-CNN, and further concatenated to gain the output ${\mathcal{V}}$:
\begin{equation}
{\mathcal{V}}= {\mathcal{F}}^{2D} \circledast {\mathcal{F}}^{3D}
\label{eq1}
\end{equation}
where $\circledast$ means concatenating.

Referring to the Gumbel-Softmax~\cite{jang2016categorical}, features during forward pass and backward pass are gained:
\begin{equation}
\begin{aligned}
{\mathcal{V}}^{\theta^{forward}_t}&={\mathcal{V}} \circ ArgMax\left( {\mathcal{AT}}, \eta_{g} \right),  \\
{\mathcal{V}}^{\theta^{backward}_t}&={\mathcal{V}} \circ SoftMax\left( {\mathcal{AT}}, \gamma \eta_{g} \right).
\label{eq2}
\end{aligned}
\end{equation}

where $\circ$ means inner product. ${\mathcal{AT}}$ is the visual attention feature map obtained by using transformer during visual encoding. $\eta_{g}$ indicates the gumbel noise which is introduced from i.i.d. random variable of $Gumbel(0, 1)$. $\gamma$ is a weighted parameter for balancing the strength of SoftMax.

Moreover, dropout helps the network to memorize the patterns of each layer evenly in each neuron, which is equivalent to increasing the robustness of the model. When the subsequent task destroys a small part of the neurons, it does not affect the overall output. Therefore, adding dropout can alleviate the catastrophic forgetting~\cite{2013An}. Specifically, to apply the structured dropout to the output of LSTM $o_{LSTM}$, we use an elementwise multiplicative mask~\cite{srivastava2014dropout, zoph2018learning} $R$ which is a randomly generated binary mask of the same shape as $o_{LSTM}$. Then we scale the elements in $R$ and replace $o_{LSTM}$ with:
\begin{equation}
D_{s}= o_{LSTM} \circ \lbrack R {{size (R)} \over {sum (R)}} \rbrack
\label{eq3}
\end{equation}
Performing the beam search~\cite{2016Diverse} on the output $D_{s}$ obtained after this step gives us the final description $\mathbf{W}$ we expect to generate.

\begin{figure}[htbp]
\centerline{\includegraphics[scale=0.25]{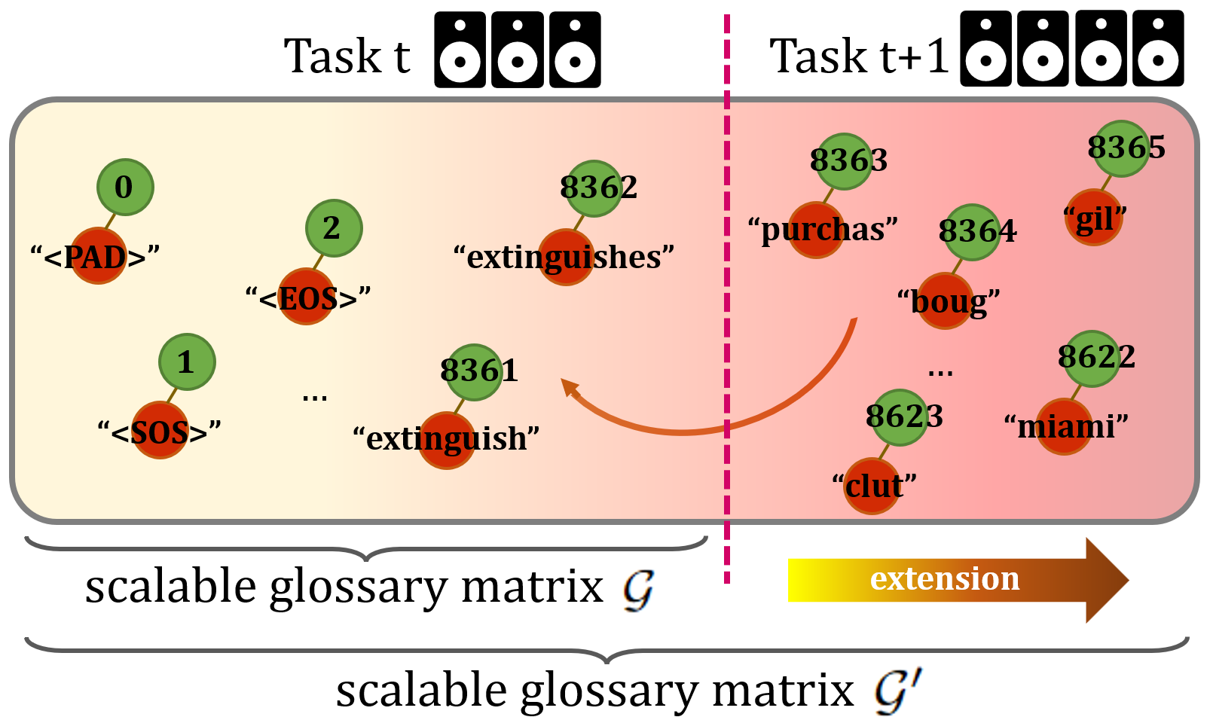}}
\vspace{-0.2cm} 
\caption{The diagram of the scalable glossary matrix in the sequential tasks. As task t+1 is entered, the glossary $\mathcal{G}'$ will gradually grow larger than the old one $\mathcal{G}$.}
\vspace{-0.3cm} 
\label{fig3}
\end{figure}

\paragraph{Basics of Parameter Migration: Glossary Ensemble}

Inspired by the way the human brain remembers, it is able to store information permanently when short-term memories in synapses are converted to long-term memories. This permanently preserved memory storage is similar to large-scale language pre-training, which can provide the model with many samples of the same type as the unknown samples in advance during the training process. We use the GloVe~\cite{pennington2014glove} initialized word embedding matrix to train with the whole network. In order to ensure that the parameters trained by the previous classes can be used to extract useful knowledge when processing incremental tasks, we design Glossary Ensemble here. Unlike the typical ensembling algorithm PathNet~\cite{fernando2017pathnet}, this paper proposes to achieve ensembling by implicitly increasing glossary. Textual information from each sequential task is added to the scalable glossary matrix. Specifically in Figure~\ref{fig3}, the glossary matrix training on the \textbf{old} task is $\mathcal{G}$. After training in the next incremental task \textbf{new}, this matrix becomes $\mathcal{G}'$. By storing this scalable glossary matrix, our network builds a segment of long-term memory units. This operation ensures that the sequence numbers corresponding to the output word matrix dimensions remain consistent during the sequential training process, which not only facilitates the memorization of textual information, but also makes it possible for the vocabulary corresponding to the parameter dimensions for the subsequent training not to change due to the order of the vocabulary lists for the new task.

\subsection{Fine-grained Sensitivity Selection of Parameters (FgSS)}

\begin{figure}[htbp]
\centerline{\includegraphics[scale=0.4]{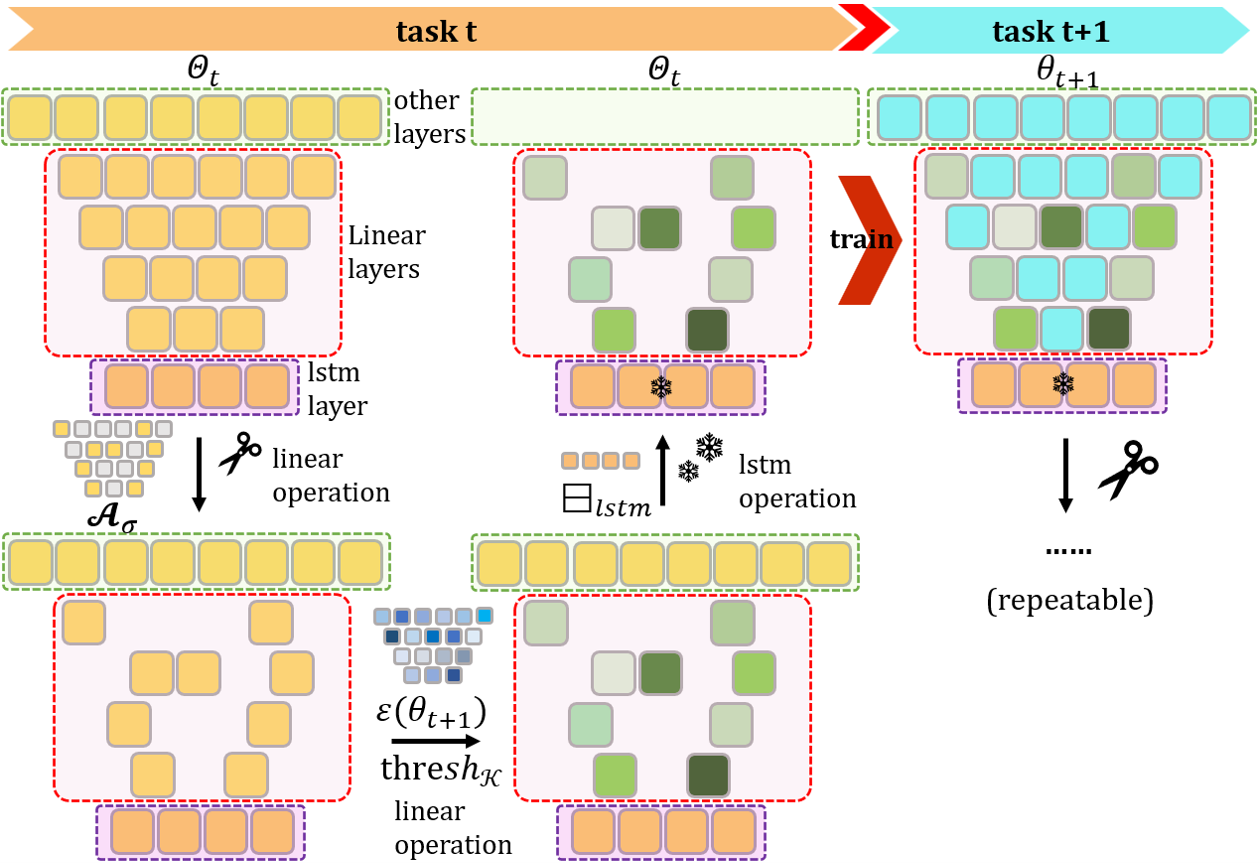}}
\caption{This schematic indicates how the FgSS module handles model parameters.
}
\vspace{-0.3cm} 
\label{fig4}
\end{figure}

We design a selection mechanism that performs fine-grained clipping of the stored old weights, reducing the computational amount of the training and limiting the performance loss of the old task. As a progressive training method, we aim to find free parameters through the proposed strategy to trade off between old and new tasks without increasing the capacity of the additional network. After the new task $t+1$ input, the retained old network parameters $\Theta_{t}$ is mapped by the technique, which is similar to the network pruning, to the corresponding layer of typical memory for subsequent training. After investigating some pruning methods~\cite{mallya2018packnet}, we find that the response of the filter changes with its sparsity and is no longer separable after passing some nonlinear layers. Therefore, the pruning ratio of each sub-task is not synchronous, and the parameters of the linear layer is mainly pruned. In detail, we randomly generate a mask $\mathcal{A}_{\sigma} \in [0, 1]$ based on the artificially set pruning ratio $\sigma$. This mask sets the pruned parameter position to 0. 

Inspired by the fisher information matrix~\cite{kirkpatrick2017overcoming}, we design the mask of linear parameters here to selectively inherit some typical parameterized information in the old network parameters. Unlike the previous method, they directly use this matrix as the weight to calculate the gap of the parameter gradient between old and new models, which is used as the incremental loss. Assuming that $n$ observations are made, this elastic weight mask for each parameter is obtained by calculating the square of the gradient:

\begin{equation}
\mathcal{E}(\theta_{t+1})=\frac{1}{\mathcal{N}} \sum_{n}\left(\frac{\partial{log(\theta_{t+1}|n)}}{\partial\theta_{t+1}}|_{\Theta_{t}}\right)^2 
\label{eq5}
\end{equation}

where $log$ means log likely-hood, $\theta_{t+1}$ is the parameter of the new network, the square term is the parameter corresponding gradient, and $\mathcal{N}$ represents the sample size.

We find that the structure of the entire video captioning backbone is shown by a branch in Figure~\ref{fig2}. Specifically, there are basic structures such as $Encoder$, $Dropout$, $Attention$, and $LSTM$. Based on the characteristics of sequence, we find that RNN forgets a lot of previous knowledge during continuous training. Therefore, we take the method of freezing the LSTM layer $\boxminus_{lstm}$ to slow down this forgetting.

These three parts make up the final design. We use a threshold function to combine the randomly linear pruned mask $\mathcal{A}_{\sigma}$ and the gradient-based elastic weight mask $\mathcal{E}(\theta_{t+1})$. Next, we attach the frozen layer $\boxminus_{lstm}$ to form the complete FgSS which is shown in Figure~\ref{fig4}:

\begin{equation}
\theta^{FgSS}_{t+1}=thresh_\kappa [\mathcal{A}_{\sigma} \circledcirc \mathcal{E}(\theta_{t+1})] \circledcirc \boxminus_{lstm}
\label{eq6}
\end{equation}

where $\kappa$ refers to the bound value chosen by the threshold function and $\circledcirc$ denotes hadamard product.

\subsection{Two-stage Knowledge Distillation (TsKD)}

To better slow down the forgetting of old knowledge, we distill the output of two layers in the backbone network. Unlike simple two-norm normalization ~\cite{hou2019learning}, this two-stage knowledge distillation can maximize the migration of knowledge from the old network to the new model. Since video captioning network training needs to fuse information in both visual and text modalities, it is important to introduce a phased cross-fusion feature and constrain it. We obtain two phased output features ($S'_A$ and $S_A$ which is shown in Figure~\ref{fig2}) in the semantic attention feature map extraction that aligns the input video with its relevant text. 

\begin{equation}
S_A= \sum_{i}\psi^{positive}_{i,j}({\mathcal{V}}_i, {\mathcal{P}}_j)
\label{eq7}
\end{equation}

where $\psi^{positive}_{i,j}$ represents the feature map from positive image-text pairs $\{i,j\}$, while ${\mathcal{V}}_i$ and ${\mathcal{P}}_j$ are the $i_{th}$ visual feature and the $j_{th}$ phrase feature separately. 

These two features contain a wealth of stylized~\cite{gatys2016image} information. In order to migrate the old style to the new features, we first design a style filter to capture the feature space of cross-modal texture information,

\begin{equation}
{S}_{cross}=\frac{\bar S_A \circ {\bar S_A}^T}{size(\bar S_A)}
\label{eq8}
\end{equation}

where $\bar S_A$ represents the row mean of the cross-feature map. Both $S'_A$ and $S_A$ will be brought into this equation for calculation.

Next, it is necessary to migrate the old style representation to the new style feature space. We minimize the mean square distance between the generated new feature map ${S}_{cross}$ and the captured original feature map ${S'}_{cross}$.

\begin{equation}
\mathcal{L}_{S_A}^{style}=\frac{1}{\mathcal{N}} \sum_{n}\left({S}_{cross}-{S'}_{cross}\right)^{2} 
\label{eq9}
\end{equation}

In order for the generated descriptive text features to inherit the knowledge of the old output, we design a tool to compare the two and optimize them. Inspired by NT-Xent loss~\cite{chen2020simple} in self-supervised learning, we first sample batch size $1 bs$ text bars for each of the old and new task output feature $\mathbf{W}_{t}, \mathbf{W}_{t+1}$, and then let the samples of each new task $z_{t+1}$ look for the old task samples $z_{t}$ corresponding to themselves during the training process, while the other $2bs-1$ samples are treated as negative samples. The similarity between the new task output with the positive sample and the remaining samples is calculated separately, aiming to aggregate the distance between the new output and the positive sample distribution while moving away from other negative samples.

\begin{equation}
\mathcal{L}_{(t+1, t)}^C=-log\frac{exp(sim(z_{t+1}, z_{t})/\tau)}{\sum^{2bs}_{g=1}\mathbbm{1}_{[g\neq t]}exp(sim(z_{t+1}, z_{g})/\tau)}
\label{eq10}
\end{equation}

where $\mathbbm{1}$ represents the indicator function, $sim(*)$ is the calculation of cosine similarity matrix, $\tau$ is the parameter for adjusting the temperature. 

During video captioning, the new task text output feature $W_{t+1}$ is optimized by the following cross entropy loss function:
\begin{equation}
\mathcal{L}_{vc}=\mathcal{L}^{CE}_{gt}(\mathbf{W}_{t+1})=-\sum\limits{log \mathbf{W}_{t+1}\left(\mathbf{W}_{t+1}^*\right)}
\label{eq4}
\end{equation}
where $\mathbf{W}_{t+1}^*$ indicates the ground truth of the new task.

The total loss of the final model training can be summed by the cross-entropy loss $\mathcal{L}_{vc}$ and the loss of two-stage knowledge distillation:

\begin{equation}
\mathcal{L}_{total}=\mathcal{L}_{vc}+\varsigma(\vartheta\mathcal{L}_{S_A}^{style} +\mathcal{L}_{(t+1, t)}^C)
\label{eq11}
\end{equation}

where $\varsigma$ and $\vartheta$ are hyperparameters used to balance between the three losses.
\section{Experimental Results}

In this section, we first segment the challenging video captioning benchmark dataset MSR-VTT~\cite{xu2016msr} following the concept of class-incremental learning. Then, we compare the proposed method with some existing methods that are portable to video captioning on some public datasets. Based on the presented experimental setup and some implementation details, the final results demonstrate the effectiveness of our model on both the four metric scores used for video captioning and the metrics we devised to detect forgetfulness. In addition, we have completed several ablation experiments to illustrate the feasibility and usefulness of the proposed new module.

\subsection{Experimental Setup}

\paragraph{Metrics}
The evaluation mainly follows the widely used official metrics in captioning, BLEU~\cite{papineni2002bleu}, ROUGE-L~\cite{lin2004rouge}, METEOR~\cite{banerjee2005meteor}, and CIDEr~\cite{vedantam2015cider}. 
With the assistant of the Microsoft COCO toolkit~\cite{chen2015microsoft}, our network achieves the final results according to the standards.

In this paper, we devise a new metric to characterize the forgetting rate of the incremental video captioning task. $\mathcal{CIDER}_{t,k}$ denotes the CIDEr of the task g after learning task t $(g\le t)$, which provides precise information about the sequential input process. We assume that the current task g contains $z^g$ classes, and the total number of all the classes that have seen (from task 1 to t) is $z^{total}$. In order to compare the specific accuracy of the current task t corresponding to each training step as well as the previous task, we designed the average step accuracy $\widetilde{\mathcal{CIDER}}_t$ at task t.

\begin{equation}
\widetilde{\mathcal{CIDER}}_t=\frac{1}{t}\sum^{t}_{g=1}\frac{z^g}{z^{total}}\mathcal{CIDER}_{t,g}
\label{eq12}
\end{equation}


\paragraph{Datasets}

MSR-VTT~\cite{xu2016msr} is a widely used benchmark for video captioning. It contains 10k open-domain videos with an average video length of 14.8 s and 20 English annotations for each clip. The content is divided into 20 simply defined categories, including travel, animation, TV shows, etc. These category labels greatly facilitate sequential segmentation tasks. Following the concept of class-incremental learning, we divide the entire dataset into 6 tasks. 
Due to the complexity of the video captioning, 10 categories are included in the base task. The remaining 10 categories are an incremental task every two. The amount of data corresponding to each label is different, and the performance displayed on the video captioning network based on transformer is quite different. Therefore, we conducted extensive experiments, and the final classification guarantees approximately similar performance for each sub-task. 
Following a common standard~\cite{zhao2019cascade, cai2023lightgcl}, we divide every sub-task according to 7 (training):1 (validation):2 (testing).

\paragraph{Implementation Details}

In the feature extraction phase, we sample features from $\ell = 20$ frames and clips per video as visual input through pre-trained 3D-CNN~\cite{hara2018can} and 2D-CNN~\cite{he2016deep, xie2017aggregated} backbones. The Adam Optimizer~\cite{kingma2014adam} is used in our pipeline with a fixed learning rate of $8.75e^{-5}$. The batch size $bs$ of each training phase is set to 8 during 16 epochs. 
There are some hyperparameters in the experiment, $\sigma$, $\kappa$, $\tau$, $\varsigma$ and $\vartheta$ are set to $50\%$, $0.01$, $0.5$, $0.6$ and $0.0001$, respectively. We generate the finial prediction statements by beam search decoding with a size of 5. For fair comparison, the training environment and backbone of all methods are the same. All the experiments used Pytorch are carried out on eight NVIDIA TITAN Xp GPUs.

\subsection{Quantitative Evaluation}

The quantitative experimental part shows the comparison of the method in this paper with some other methods and the use effect of some important modules.

\begin{table}[htbp]
\caption{The results of our method and two typical knowledge distillation adding Glossary Ensemble are quantitatively compared using four metrics ($\%$), where the value in bold is the maximum value on MSR-VTT.}
\begin{adjustbox}{center}
\scalebox{0.7}{
\begin{tabular}{c|c|ccccc}
\hline
\multicolumn{1}{c|}{\multirow{2}{*}{Model}}&\multicolumn{1}{c|}{\multirow{2}{*}{Train Step}}&\multicolumn{5}{c}{Performance}\\
  & &\textit{BLEU@4$\uparrow$}& \textit{METEOR$\uparrow$}& \textit{ROUGE-L$\uparrow$}& \textit{CIDEr$\uparrow$}& \textit{$\widetilde{\mathcal{CIDER}}_t\uparrow$}\\
\hline\hline
\multirow{6}{*}{Fine-tune}& Base & 35.91& 26.21& 58.03 & 35.33 & 35.33\\
  & +CIL1 & 27.45& 21.75& 52.14 & 15.99 & 4.59\\
  & +CIL2 & 26.82& 20.83& 50.33 & 17.99 & 2.63\\
  & +CIL3 & 25.17& 20.56& 51.15 & 16.25 & 1.14\\
  & +CIL4 & 30.11& 22.48& 53.49 & 22.80 & 0.96\\
  & +CIL5 & 27.16& 20.91& 51.38 & 18.62 & 0.81\\
\hline \hline
\multirow{6}{*}{Ideal$^{\dag}$}& Base & 35.91& 26.21& 58.03 & 35.33 & 35.33\\
  & +CIL1 & 35.30& 26.36& 57.65 & 35.07& 18.07\\
  & +CIL2 & 35.73& 26.21& 57.42 & 36.83 & 12.51\\
  & +CIL3 & 36.61& 26.08& 57.54 & 38.58 & 8.95\\
  & +CIL4 & 37.17& 26.52& 58.31 & 38.67 & 7.06\\
  & +CIL5 & 37.34& 26.64& 58.37 & 38.83 & 6.34\\
\hline \hline
\multirow{6}{*}{$LwF$}& Base & 35.91& 26.21& 58.03 & 35.33 & 35.33\\
  & +CIL1 & 29.17& 24.97& 54.02 & 25.57&13.42\\
  & +CIL2 & 26.93& 24.14& 52.36 & 25.68&8.71\\
  & +CIL3 & 23.57& 22.44& 49.73 & 23.68&5.85\\
  & +CIL4 & 27.93& 23.71& 51.86 & 26.78&5.32\\
  & +CIL5 & 24.25& 22.23& 49.11 & 22.08&3.89\\
 \hline
\multirow{6}{*}{$EWC$}& Base & 35.91& 26.21& 58.03 & 35.33 & 35.33\\
  & +CIL1 & 30.70& 23.81& 53.91 & 26.39&12.73\\
  & +CIL2 & 30.89& 23.97& 53.75 & 27.78&9.46\\
  & +CIL3 & 32.31& 23.64& 55.02 & 27.07&6.62\\
  & +CIL4 & 32.73& 24.52& 55.51 & 28.98&5.53\\
  & +CIL5 & 29.97& 22.31& 52 61 & 23.58&3.89\\
\hline
\textbf{\multirow{6}{*}{MCF-VC}}& Base & \textbf{35.91}& \textbf{26.21}& \textbf{58.03} & \textbf{35.33} & \textbf{35.33}\\
  & +CIL1 & \textbf{33.66}& \textbf{25.36}& \textbf{55.90} & \textbf{30.59}& \textbf{15.52}\\
  & +CIL2 & \textbf{34.11}& \textbf{24.67}& \textbf{55.36} & \textbf{31.21}& \textbf{10.37}\\
  & +CIL3 & \textbf{33.99}& \textbf{24.68}& \textbf{56.25} & \textbf{31.52}& \textbf{7.68}\\
  & +CIL4 & \textbf{34.48}& \textbf{25.01}& \textbf{56.66} & \textbf{30.53} & \textbf{5.91}\\
  & +CIL5 &  \textbf{31.42}& \textbf{23.10}& \textbf{53.46} & \textbf{26.57}&  \textbf{4.25}\\
\hline
\multicolumn{7}{l}{\scriptsize{$^{\dag}$old \& new sub-tasks are visible in training step according to our segmentation which is different from standard}}
\end{tabular}}
\end{adjustbox}
\vspace{-0.5cm} 
\label{tab3}
\end{table}

\begin{figure}[htbp]
\centerline{\includegraphics[scale=0.4]{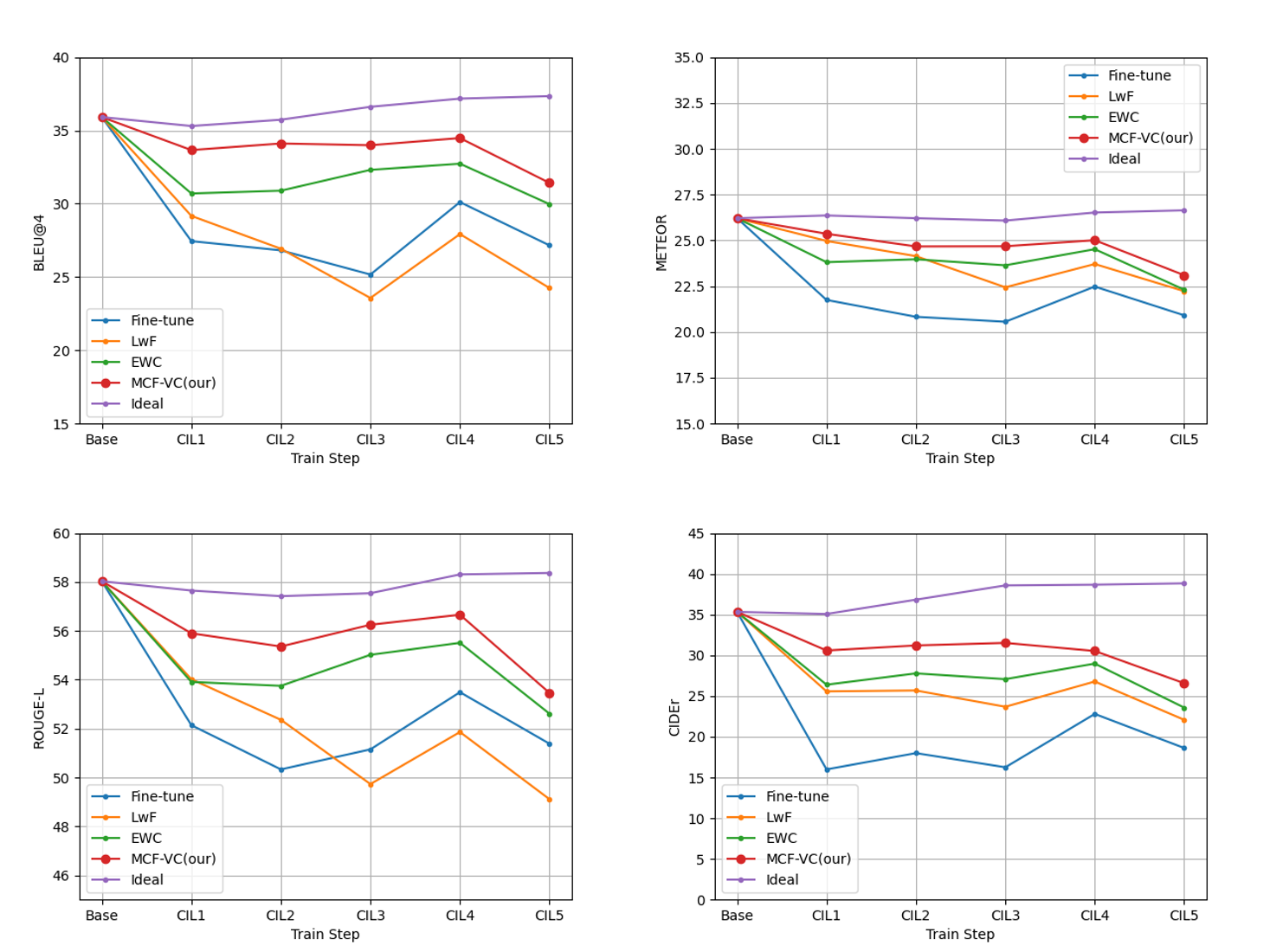}}
\vspace{-0.2cm} 
\caption{In order to better evaluate the significant performance improvement of the incremental approach proposed in this paper compared to other approaches on martics in natural language processing, we use line graphs to clearly express it.}
\vspace{-0.5cm} 
\label{fig7}
\end{figure}

\paragraph{Comparison with Ideal Performance}

We evaluate the performance of our proposed method MCF-VC on MSR-VTT benchmark divided according to class-incremental task. It is reported along with two other methods: simple fine-tuning(denoted by ``Fine-tune") and train the model with all old and new tasks visible (denoted by ``Idea"). In this article, we think of ``Fine-tune" as the lower bound and ``Idea" as the upper bound(as shown in Table~\ref{tab3}). It is clear that the direct fine-tuning is much lower than the ideal value, which further corroborates the existence of stability-plasticity in the video captioning with sequential inputs.

\paragraph{Comparison with State-Of-The-Art}

Table~\ref{tab3} also summarizes the performance of our proposed method and compares it with class-incremental methods on other standard image classification tasks, including LwF~\cite{li2017Learning} and EWC~\cite{kirkpatrick2017overcoming}. After training on each new task step, we evaluate the performance on a test set containing both old and new tasks. LwF is similar to joint training, which does not use the data and labels of the old task and only needs to use the data of the new task, and optimizes the new task to improve the accuracy of model prediction on the new task and maintain the prediction performance of the neural network for the previous task. EWC uses the fisher information matrix to constrain the network parameters and reduce the degree of forgetting of previous knowledge by the model. 

Considering the particularity of the backbone method video captioning, the replay method for storing old samples is not suitable. Because this kind of storage not only does not conform to the real scene, but also has storage difficulties and video copyright problems. The method of dynamic network expansion is mainly suitable for networks with multiple classification heads in classification problems, and the solution of such problems has a huge semantic gap with the video captioning, so it cannot be directly ported for comparison. Additionally, since the consistency of vocabulary records has a significant impact on the size of model parameters, this paper selects two class-incremental learning works in image classification adding Glossary Ensemble for comparison in this section.

\begin{figure*}[htbp]
\centerline{\includegraphics[scale=0.6]{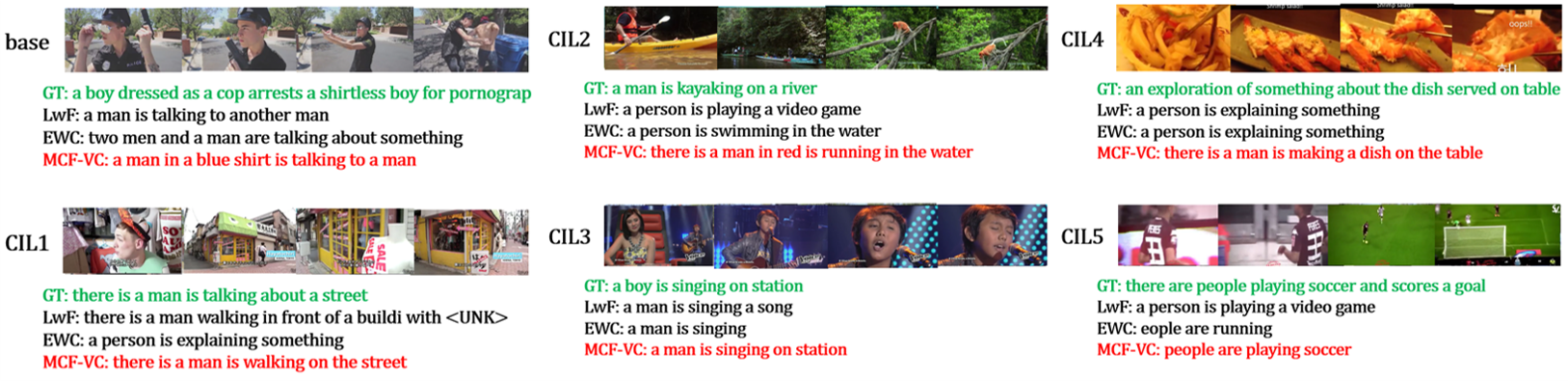}}
\vspace{-0.3cm} 
\caption{Qualitative results on 6 videos from base to CIL1 sub-tasks of MSR-VTT in the last incremental duration. ``GT" is one of the descriptions in ground-truth. While, ``LwF" and ``EWC" represent the output captions of two typical class-increnmental methods. Comparing the three sentences, it is found that the output of our network is effectively applied to the classincremental setting task of sequential input, and maintains the good performance of the old task while learning the new task.}
\vspace{-0.3cm} 
\label{fig5}
\end{figure*}

As shown in Table~\ref{tab3} and Figure~\ref{fig7}, we observe that our method is significantly superior to other distillation-based methods. This confirms that MCF-VC can effectively solve the catastrophic forgetting of sequential inputs in video captioning without storing old training samples or dynamically increasing the model structure. After 5 incremental tasks on the MSR-VTT baseline, the difference between MCF-VC and LwF is $3.75-7.84\%$ and $1.55-4.45\%$ with EWC. The proposed approach achieves performance advantages over these state-of-the-art incremental methods.

\begin{table}[htbp]
\vspace{-0.2cm} 
\caption{Ablation studies are carried out on the modules ($\mathcal{G}$, FgSS and TsKD) proposed by MCF-VC on MSR-VTT.}
\vspace{-0.2cm} 
\begin{adjustbox}{center}
\scalebox{0.8}{
\begin{tabular}{ccc|c|cccc}
\hline
\multicolumn{3}{c|}{Module}&\multicolumn{1}{c|}{\multirow{2}{*}{Sub-task}}&\multicolumn{4}{c}{Performance}\\
\textbf{$\mathcal{G}$} &\textbf{FgSS} &\textbf{TsKD} & &\textit{BLEU@4}& \textit{METEOR}& \textit{ROUGE-L}& \textit{CIDEr}\\
\hline
\multirow{6}{*}{\textbf{$\times$}}&\multirow{6}{*}{$\times$}& \multirow{6}{*}{$\times$} & Base & 35.91& 26.21& 58.03 & 35.33\\
 &&  & +CIL1 & 27.45& 21.75& 52.14 & 15.99\\
  && & +CIL2 & 26.82& 20.83& 50.33 & 17.99\\
  && & +CIL3 & 25.17& 20.56& 51.15 & 16.25\\
  && & +CIL4 & 30.11& 22.48& 53.49 & 22.80\\
 &&  & +CIL5 & 27.16& 20.91& 51.38 & 18.62\\
 \hline
\multirow{6}{*}{\textbf{$\surd$}}&\multirow{6}{*}{$\times$}& \multirow{6}{*}{$\times$} & Base & 35.91& 26.21& 58.03 & 35.33\\
  && & +CIL1 & 29.16& 22.91& 52.27 & 22.75\\
  && & +CIL2 & 27.87& 23.02& 52.59 & 24.28\\
  && & +CIL3 & 26.57& 21.85& 52.44 & 22.58\\
  && & +CIL4 & 29.48& 23.19& 53.58 & 26.04\\
 && & +CIL5 & 27.52& 21.55& 51.03 & 21.82\\
 \hline
\multirow{6}{*}{\textbf{$\surd$}}&\multirow{6}{*}{\textbf{$\surd$}}& \multirow{6}{*}{$\times$} & Base & 35.91& 26.21& 58.03 & 35.33\\
 & & & +CIL1 & 33.59& 25.20& 55.81 & 29.98\\
 &&  & +CIL2 & 33.76& 24.32& 55.71 & 31.04\\
 &&  & +CIL3 & 33.54& 24.20& 55.77 & 29.99\\
 &&  & +CIL4 & 34.34& 25.06& 56。45 & 30.12\\
 &&  & +CIL5 & 31.05& 23.21& 53.61 & 25.54\\
 \hline
\multirow{6}{*}{\textbf{$\surd$}}&\multirow{6}{*}{\textbf{$\times$}}& \multirow{6}{*}{\textbf{$\surd$}} & Base & 35.91& 26.21& 58.03 & 35.33\\
& &  & +CIL1 & 32.85& 24.79& 55.04 & 29.35\\
& &  & +CIL2 & 31.35& 24.03& 54.67 & 28.96\\
& &  & +CIL3 & 31.31& 23.67& 55.46 & 27.75\\
& &  & +CIL4 & 33.38& 24.70& 55.75 & 29.74\\
& &  & +CIL5 & 31.40& 23.03& 53.68 & 26.52\\
\hline
\end{tabular}}
\end{adjustbox}
\label{tab4}
\vspace{-0.5cm} 
\end{table}

\paragraph{Ablation Study}

Here, we analyze the effects of several important modules presented in isolation. As can be seen from the results of Table~\ref{tab4}: (1) Adding Glossary Ensemble to the backbone network is effective in mitigating forgetfulness. (2) In networks without FgSS and TsKD, distillation using only the simplest old and new models (as the second line) is a complete failure. (3) FgSS successfully picks out old knowledge for inheritance and achieved better results, which improves the performance of simple distillation on MSR-VTT by $2.99-7.09\%$. (4) Baseline combined with TsKD significantly improves the original performance, which increases the performance range by $3.2-6.46\%$. In addition, it is reported that the effectiveness of TsKD is more pronounced with the help of FgSS, which shows that TsKD and FgSS can benefit each other.





\section{Qualitative Evaluation}

We show the qualitative results of the proposed method on the MSR-VTT dataset in Figure~\ref{fig5}, including key frames, ground truth of captions, results of the existing continuous learning methods using the same backbone and modified strategies, and the captions predicted by our model. As shown in Figure~\ref{fig5}, we finish training the last incremental sub-task and test all sub-tasks. Compared to the output descriptions of ground-truth, LwF, and EWC, our method performs best in the case of five incremental tasks. Here, we show some qualitative plots to illustrate the final effect achieved by our model.
\section{Conclusions}

In this paper, we propose a method to Mitigate Catastrophic Forgetting in class-incremental learning for multimodal Video Captioning (MCF-VC). This is the first proposed solution to the model forgetting problem resulting from progressive training in the video captioning. In order to better adapt to class-incremental learning, we have carefully modified the transformer-based encoder-decoder architecture to add some uniquely designed modules. At the same time, we design a Fine-grained Sensitivity Selection (FgSS) to adaptively select the knowledge of the old model with positive correlation with the new task. Inheriting the old knowledge at the level of detail and then training on the global basis of the new task, we can ultimately make a relatively best trade-off between the old and new tasks. Next, a Two-stage Knowledge Distillation (TsKD) is proposed to constrain some of the phased features of the old and new tasks. It is divided into two phases, with different types of distillation on the visual text cross modal feature and the final text output feature, respectively. It can inherit the style information of the old features well and migrate to the new features, and finally present an excellent performance on both the old and new tasks. Experiments on the MSR-VTT dataset show that our method achieves a significant performance improvement compared to other classical class incremental learning methods, especially almost $1.10\%$ improvement on the $\widetilde{\mathcal{CIDER}}_t$. This illustrates that our method solves the catastrophic forgetting caused by sequential input of videos on the video captioning, which can effectively alleviate the forgetting of old knowledge while learning new tasks.
{
    \small
    \bibliographystyle{ieeenat_fullname}
    \bibliography{main}
}

\clearpage
\setcounter{page}{1}
\maketitlesupplementary


Due to space limitation of the paper, we present some supporting experiments and descriptions in the appendix.

\subsection{Experimental Setup}

\paragraph{Metrics}
BLEU-n~\cite{papineni2002bleu} focuses primarily on n-grams overlapping between predictive text and ground truth. METEOR~\cite{banerjee2005meteor} and ROUGE-L~\cite{lin2004rouge} mainly calculate how many n-grams from ground truth are generated captions, that is, recall. CIDEr~\cite{vedantam2015cider}, as the most important indicator, assigns greater weight to infrequent phrases through TF-IDF vectors.

\paragraph{Datasets}

MSVD~\cite{chen2011collecting} is a video captioning benchmark containing 1970 YouTube open-domain video clips. The length of these short videos ranged from 10s to 25s, with an average length of 9s.  Its annotations are multilingual, and we only choose English annotations. According to some previous studies, the whole dataset can be divided into a training set (1200), a validation set (100) and a test set (670). We choose it to show the domain migration nature of the model introduced in this paper.

\paragraph{Implementation Details}
Since the experiments in this paper are presented for the first time, the design details of the experiments are listed here. The test dataset corresponding to each incremental step in the paper is the relationship shown in Table~\ref{tab10}.

\begin{table}[htbp]
\caption{The class-incremental experiments based on the MSR-VTT dataset are correspondingly written in the main text to record the names of the sub-tasks that are added, while tests is performed on the full learned data after each incremental training step.}
\begin{adjustbox}{center}
\scalebox{0.8}{
\begin{tabular}{c|c|c}
\hline
\multicolumn{2}{c|}{Steps}&\multicolumn{1}{c}{\multirow{2}{*}{Test sub-tasks}}\\
\cline{1-2}
Train Step & Sub-task & \\
\hline
Base& Base & Base\\
+CIL1  & +CIL1 & Base+CIL1\\
+CIL2  & +CIL2 & Base+CIL1+CIL2\\
+CIL3  & +CIL3 & Base+CIL1+CIL2+CIL3\\
+CIL4  & +CIL4 & Base+CIL1+CIL2+CIL3+CIL4\\
+CIL5  & +CIL5 & Base+CIL1+CIL2+CIL3+CIL4+CIL5\\
 \hline
\end{tabular}}
\end{adjustbox}
\label{tab10}
\end{table}

\section{Quantitative Evaluation}

\subsection{Incremental Domain Migration}

To verify the existence of forgetting, we conduct complementary experiments on publicly available datasets from two different domains. We train the tested model for migration across domains on the new MSR-VTT data after learning it on the MSVD dataset first. The final model obtained is tested on the two datasets separately and the results are shown in Table~\ref{tab11}. It can be found that the traditional video captioning method performs poorly on the old data MSVD by learning the MSVD data first and then supplementing it with the MSR-VTT data. Compared to other incremental learning methods, our proposed method has a good performance on both old and new datasets.

\subsection{Comparison with Traditional Video Captioning}
In this paper, the MSR-VTT dataset is divided in a different ratio than the official division, specifically the official one is split according to $65 \%:30 \%:5 \%$. In order to get the performance on the new segmented dataset, we need to do a new set of experiments. With the newly split training:validation:test ratio as shown in Table~\ref{tab5}, we perform a series of tests on the baseline~\cite{ryu2021semantic} and our proposed architecture. Obviously, our proposed method outperforms the traditional video captioning method under the training condition where all data is fixedly visible.

\begin{table}[htbp]
\caption{Our method MCF-VC compares with baseline SGN~\cite{ryu2021semantic} on newly divided MSR-VTT.}
\begin{adjustbox}{center}
\scalebox{0.9}{
\begin{tabular}{c|cccc}
\hline
\multicolumn{1}{c|}{\multirow{2}{*}{Model}}&\multicolumn{4}{c}{\textbf{MSR-VTT} - 7 (train):1 (valid):2 (test)}\\
\cline{2-5}
 & \textit{BLEU@4}& \textit{METEOR}& \textit{ROUGE-L}& \textit{CIDEr}\\
\hline
SGN~\cite{ryu2021semantic}  & 36.11& 26.28& 58.08 & 38.51\\
\textbf{MCF-VC}  & \textbf{37.34}& \textbf{26.64}& \textbf{58.37} & \textbf{38.83}\\
\hline
\end{tabular}}
\end{adjustbox}
\label{tab5}
\end{table}

\subsection{Calculation of our designed metric}

\begin{figure}[htbp]
\vspace{-0.2cm} 
\centerline{\includegraphics[scale=0.56]{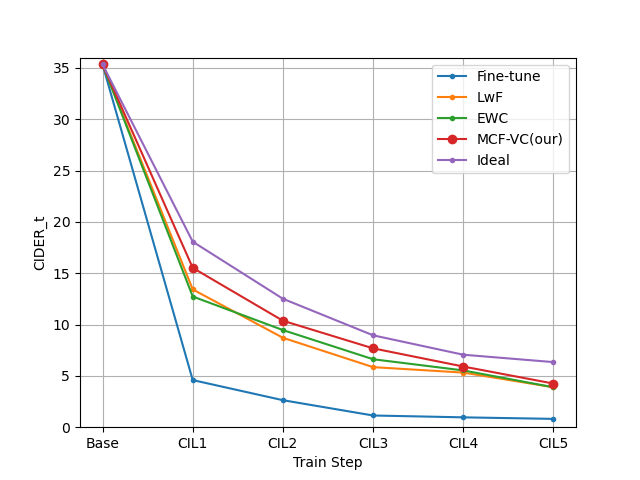}}
\caption{Compared to the LwF, EWC method, our experiment performed best on five incremental tasks. Here, the line chart records the forgetting curves for these methods.
}
\vspace{-0.2cm} 
\label{fig6}
\end{figure}

To compute the $\widetilde{\mathcal{CIDER}}_t$ of our design, we test each sub-task after the category increment step corresponding to each MSR-VTT dataset. Our experiments record the values of four metrics widely used to assess captiong quality. Due to space limitations and typographical difficulties, only CIDEr values are listed in Table~\ref{tab12}. Based on the formulas in the paper, we calculate the corresponding $\widetilde{\mathcal{CIDER}}_t$ after each incremental step and plot a forgetting curve based on these corresponding values. As shown in Figure~\ref{fig6}, our method is more resistant to forgetting than other methods.

\begin{table*}[htbp]
\caption{The class-incremental experiments based on the MSR-VTT dataset are correspondingly written in the main text to record the names of the sub-tasks that are added, while tests is performed on the full learned data after each incremental training step.}
\begin{adjustbox}{center}
\scalebox{0.8}{
\begin{tabular}{c|cccc|cccc}
\hline
\multicolumn{1}{c|}{\multirow{2}{*}{Model}}&\multicolumn{4}{c|}{MSVD}&\multicolumn{4}{c}{MSR-VTT - 7 (train):1 (valid):2 (test)}\\
\cline{2-9}
& \textit{BLEU@4}& \textit{METEOR}& \textit{ROUGE-L}& \textit{CIDEr}& \textit{BLEU@4}& \textit{METEOR}& \textit{ROUGE-L}& \textit{CIDEr}\\
\hline
SGN~\cite{ryu2021semantic} & 12.73 & 17，76 & 45.73 & 6.35 & 35，04 & 25.20 & 56.92 & 35.30 \\
\hline
LwF~\cite{li2017Learning}&19.96 & 23.28 & 52.15 & 23.32& 4.03 & 12.56 & 31.96 & 1.97 \\
EWC~\cite{kirkpatrick2017overcoming}& 24.77&23.58&53.94& 21.16 &36.88&25.93&57.48&37.54\\
\textbf{MCF-VC (our)}& \textbf{29.21}& \textbf{26.50} & \textbf{57.65} & \textbf{28.89} & \textbf{34.09} & \textbf{25.00} & \textbf{55.98} & \textbf{31.17}\\
 \hline
\end{tabular}}
\end{adjustbox}
\label{tab11}
\end{table*}

\begin{table}[htbp]
\caption{Test results for each sub-task corresponding after the incremental step.}
\begin{adjustbox}{center}
\scalebox{0.8}{
\begin{tabular}{c|c|cccccc}
\hline
\multicolumn{1}{c|}{\multirow{2}{*}{Model}}&\multicolumn{1}{c|}{\multirow{2}{*}{Train Step}}&\multicolumn{6}{c}{Test of \textit{CIDEr}}\\
 & &Base& +CIL1& +CIL2& +CIL3& +CIL4 &+CIL5\\
\hline
\multirow{6}{*}{Fine-tune}& Base & 35.33\\
  & +CIL1 & 1.95& 45.36\\
  & +CIL2 & 0.53& 2.18&  50.48 \\
  & +CIL3 & 0.36& 1.04& 0.84 & 32.91\\
  & +CIL4 & 1.38& 0.68& 0.72 & 1.03& 33.95\\
  & +CIL5 & 1.25& 1.88& 1.47 & 0.85&0.80&37.08\\
 \hline
\multirow{6}{*}{Ideal$^{\dag}$}& Base & 35.33\\
  & +CIL1 & 35.51&39.25\\
  & +CIL2 & 33.61&41.94&52.79\\
  & +CIL3 & 33.02&43.37&47.24&30.75\\
  & +CIL4 & 34.17&39.80&51.95&21.10&33.95\\
  & +CIL5 & 35.09&44.49&54.80&31.81&33.96&40.26\\
\hline
\multirow{6}{*}{$LwF$}& Base & 35.33\\
  & +CIL1 & 26.08&30.54\\
  & +CIL2 & 22.96&29.64&38.36\\
  & +CIL3 & 21.33&30.91&38.59&10.93\\
  & +CIL4 & 25.86&32.47&40.99&14.72&22.03\\
  & +CIL5 & 25.28&31.75&40.93&15.94&23.54&14.14\\
\hline
\multirow{6}{*}{$EWC$}& Base & 35.33\\
  & +CIL1 & 22.78&38.86\\
  & +CIL2 & 23.10&31.73&51.45\\
  & +CIL3 & 22.26&29.44&41.77&29.40\\
  & +CIL4 & 24.00&34.69&49.14&14.22&30.80\\
  & +CIL5 & 17.04&31.01&45.06&15.57&23.38&33.28\\
\hline
\textbf{\multirow{6}{*}{MCF-VC}}& Base & \textbf{35.33}\\
  & +CIL1 & \textbf{29.73}& \textbf{37.60}\\
  & +CIL2 & \textbf{26.67}& \textbf{33.49}& \textbf{50.86}\\
  & +CIL3 & \textbf{26.66}& \textbf{30.00}& \textbf{49.50} & \textbf{33.11}\\
  & +CIL4 & \textbf{25.50}& \textbf{34.90}& \textbf{50.29} & \textbf{22.57}& \textbf{30.79}\\
  & +CIL5 &  \textbf{21.37}& \textbf{32.95}& \textbf{44.82} & \textbf{19.93}& \textbf{23.99}& \textbf{26.22}\\
\hline

\end{tabular}}
\end{adjustbox}
\label{tab12}
\end{table}

\subsection{Ablation Study}

For the proposed Fine-grained Sensitivity Selection of Parameters (FgSS), we perform multiple sets of experiments on the frozen parameter layer. Experiments show that the frozen RNN parameter layer method designed in this paper can better inherit the old knowledge and weigh the new knowledge. As shown in Table~\ref{tab1}, after freezing the parameter layer of RNN, the comprehensive performance improvement in all aspects is the greatest. It may be because RNNs unfold in a time series, that is, what happened in the previous moment will have an impact on the development of things in the future. This feature causes the freezing parameters to allow the model to better perform anti-forgetting learning from old and new tasks, rather than just learning from new training data.





\begin{table}[htbp]
\caption{We perform multiple sets of ablation for the frozen parameter layer in the proposed FgSS on MSR-VTT.}
\begin{adjustbox}{center}
\scalebox{0.8}{
\begin{tabular}{c|c|cccc}
\hline
\multicolumn{1}{c|}{\multirow{2}{*}{Frozen Layer}}&\multicolumn{1}{c|}{\multirow{2}{*}{Sub-task}}&\multicolumn{4}{c}{Performance}\\
 & &\textit{BLEU@4}& \textit{METEOR}& \textit{ROUGE-L}& \textit{CIDEr}\\
\hline
\multirow{6}{*}{Attention}& Base & 35.91& 26.21& 58.03 & 35.33\\
  & +CIL1 & 32.09& 24.49& 55.22 & 28.65\\
  & +CIL2 & 33.77& 24.91&  55.65&29.65 \\
  & +CIL3 & 32.29& 24.16& 55.57 & 28.89\\
  & +CIL4 & 32.81& 24.58& 55.78 & 29.20\\
  & +CIL5 & 30.52& 22.64& 53.47 & 24.46\\
 \hline
\multirow{6}{*}{Dropout}& Base & 35.91& 26.21& 58.03 & 35.33\\
  & +CIL1 & 32.76& 24.73& 54.95 & 28.80\\
  & +CIL2 & 32.76& 24.48& 55.10 & 29.70\\
  & +CIL3 & 31.65& 23.86& 55.59 & 28.01\\
  & +CIL4 & 32.72& 24.33& 55.71 & 29.08\\
  & +CIL5 & 29.82& 22.48& 53.15 & 24.13\\
\hline
\multirow{6}{*}{Encoder}& Base & 35.91& 26.21& 58.03 & 35.33\\
  & +CIL1 & 31.91& 24.441& 54.80 & 28.21\\
  & +CIL2 & 33.04& 24.46& 54.96 & 29.97\\
  & +CIL3 & 31.38& 23.69& 55.23 & 28.03\\
  & +CIL4 & 33.96& 24.57& 56.12 & 29.43\\
  & +CIL5 & 31.32& 23.26& 54.08 & 25.34\\
\hline
\textbf{\multirow{6}{*}{RNN}}& Base & \textbf{35.91}& \textbf{26.21}& \textbf{58.03} & \textbf{35.33}\\
  & +CIL1 & \textbf{33.66}& \textbf{25.36}& \textbf{55.90} & \textbf{30.59}\\
  & +CIL2 & \textbf{34.11}& \textbf{24.67}& \textbf{55.36} & \textbf{31.21}\\
  & +CIL3 & \textbf{33.99}& \textbf{24.68}& \textbf{56.25} & \textbf{31.52}\\
  & +CIL4 & \textbf{34.48}& \textbf{25.01}& \textbf{56.66} & \textbf{30.53}\\
  & +CIL5 &  \textbf{31.42}& \textbf{23.10}& \textbf{53.46} & \textbf{26.57}\\
\hline

\end{tabular}}
\end{adjustbox}
\label{tab1}
\end{table}

\end{CJK}
\end{document}